\documentclass{INTERSPEECH2023}
\usepackage{hyperref}
\usepackage[dvipsnames]{xcolor}
\usepackage{bm}
\usepackage{cleveref}
\usepackage{multirow}
\usepackage{lipsum}

\definecolor{sns_blue}{HTML}{4c72b0}
\definecolor{sns_orange}{HTML}{dd8452}
\definecolor{sns_green}{HTML}{55a868}
\definecolor{sns_red}{HTML}{c44e52}

\interspeechcameraready 

\newcommand\blfootnote[1]{%
  \begingroup
  \renewcommand\thefootnote{}\footnote{#1}%
  \addtocounter{footnote}{-1}%
  \endgroup
}

\title{Can Self-Supervised Neural Representations Pre-Trained on Human Speech distinguish Animal Callers?}

\name{Eklavya Sarkar$^{1,2}$ and Mathew Magimai.-Doss$^1$}
\address{
  $^1$Idiap Research Institute, Martigny, Switzerland\\
  $^2$Ecole polytechnique fédérale de Lausanne, Switzerland}
\email{\{eklavya.sarkar, mathew\}@idiap.ch}

\begin{document}

\maketitle
 
\begin{abstract}
Self-supervised learning (SSL) models use only the intrinsic structure of a given signal, independent of its acoustic domain, to extract essential information from the input to an embedding space. This implies that the utility of such representations is not limited to modeling human speech alone. Building on this understanding, this paper explores the cross-transferability of SSL neural representations learned from human speech to analyze bio-acoustic signals. We conduct a caller discrimination analysis and a caller detection study on Marmoset vocalizations using eleven SSL models pre-trained with various pretext tasks. The results show that the embedding spaces carry meaningful caller information and can successfully distinguish the individual identities of Marmoset callers without fine-tuning. This demonstrates that representations pre-trained on human speech can be effectively applied to the bio-acoustics domain, providing valuable insights for future investigations in this field.
\end{abstract}

\section{Introduction} \label{sec:intro}
\blfootnote{Code: \href{https://github.com/idiap/ssl-caller-detection}{\url{github.com/idiap/ssl-caller-detection}}}
The study of animal vocalizations, or bio-acoustics, has progressed significantly in recent years due to approaches inherited from machine learning and deep learning \cite{PMID:35341043}. However, most of these are supervised approaches, which require large amounts of labeled data, which is often scarce in bio-acoustics. Self-supervised representation learning (SSL) has emerged as a powerful tool in speech processing to leverage unlabeled data by pre-training models to solve pretext tasks using surrogate labels created from the structure inherent to the data itself. Given an acoustic waveform signal as input, an SSL model uses said labels and the pretext task to train and iteratively optimize its learning objective. The information encoded in the representations can vary depending on the selected learning objective, which can be roughly categorized into generative and discriminative approaches. Generative methods try to either reconstruct masked acoustic frames \cite{liu2020nonautoregressive, 9054458, 10.1109/TASLP.2021.3095662}, or predict future frames using an auto-regressive framework \cite{chung2019unsupervised, chung2020vqapc}. Discriminative approaches either learn by contrastive learning, i.e. discriminating positive samples from negative ones \cite{riviere2020unsupervised, baevski2020wav2vec}, or else by predicting pseudo-labels of discrete masked regions \cite{hsu2021hubert, 9814838, pmlr-v162-baevski22a} or the output of specific hidden layers \cite{chang2022distilhubert}. The representations learnt from the chosen SSL model can then be further fine-tuned to a wide range of speech downstream tasks, which have yielded state-of-the-art results on the SUPERB benchmark \cite{yang21c_interspeech}.

Self-supervised learning only utilizes the intrinsic structure of unlabeled data without any reliance on domain-specific knowledge, such as human speech production, to capture essential information about the input data, and extract high-level representations in an embedding space. Thus, the utility of such representations may not only be restricted for modeling human speech, as demonstrated by recent works on other acoustic domains such as music \cite{9414405, zeng-etal-2021-musicbert} and biomedical signals \cite{Banville_2021, 8918693}. Given this understanding, and the fact that both humans and animals have a voice production system, our objective is to investigate the cross-transferability of representations learned from human speech for analyzing animal vocalizations.

To that end, we conduct an animal caller detection study on Marmoset (\textit{Callithrix jacchus}) vocalizations, and demonstrate its applicability through means of eleven different SSL models pre-trained with different pretext tasks. Our study also aims to provide practical benefits to biologists and ethologists by providing a framework to distinguish individual identities \textit{within} the same animal species, which is an understudied topic in bio-acoustics and a much harder problem than across-species classification \cite{PMID:35341043}. Some previous works has explored birdsong detection \cite{9413528} and bio-acoustic event detection \cite{Bermant2022.10.12.511740} using contrastive learning, however, the generalization of SSL models to animal vocalizations has largely remained unexplored. To the best of our knowledge, no previous study has looked into caller detection by utilizing the embedding space learnt by pre-training on human speech.

\section{Study Design} \label{sec:study_design}
This section presents the study design to systematically investigate the cross-transferability of representations learned from human speech for animal caller detection. Specifically, we design a study with the following research questions:
\begin{enumerate}
\item[1.] How discriminative are the embedding spaces of SSL models pre-trained on human speech? 
\item[2.] Can we systematically detect individual Marmoset callers using said embedding space?
\end{enumerate}
The remainder of the section presents the dataset, research framework, and selection of SSL models for our investigations.

\subsection{Dataset}
\begin{figure}[ht]
  \centering
  \includegraphics[width=\linewidth]{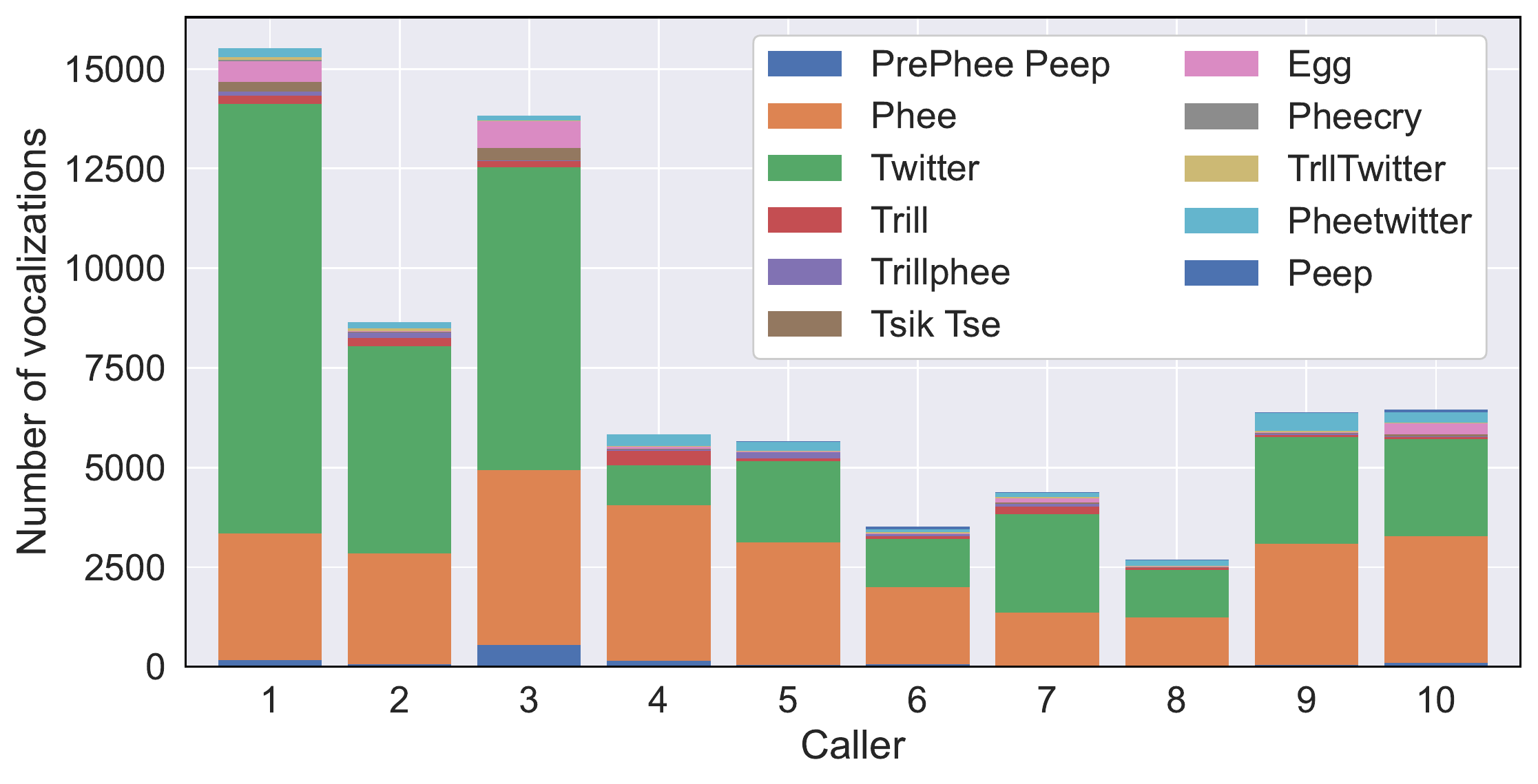}
  \caption{Vocalization per callers grouped by call-type.}
  \label{fig:data_distribution}
\end{figure}

For our study, we requested and used the marmoset dataset collected and labeled by \cite{10.1121/1.5047743}. The dataset contains audio recordings of eleven different marmoset calltypes, such as Twitters, Phees, and Trills, manually annotated using the Praat tool. The audio was recorded from five pairs of infant marmoset twins, each recorded individually in two separate sound-proofed recording rooms at a sampling rate of $44.1$ kHz. The start and end time, call type, and marmoset identity of each vocalization are provided, labeled by an experienced researcher. The data contains 350 files of precisely labelled 10-minute audio recordings across all caller classes. We downsample the data to $16$ kHz, remove all segments labeled as `silence' and `noise', and only keep the vocalization segments, amounting to a total of $464$ minutes over $72,921$ vocalization segments, with a mean and median length of $381 \pm 375$ ms and $127$ ms respectively. \Cref{fig:data_distribution} shows the imbalanced distribution of vocalizations per caller, color coded by calltype. We divide the entire data into training, validation, and test sets, named \textit{Train}, \textit{Val}, and \textit{Test} respectively, following a 70:20:10 split. This distribution allows us to train models on a sufficiently large dataset while ensuring that we have sufficient data for model evaluation and validation. \textit{Train} is used to train the models, \textit{Val} to tune hyperparameters, and \textit{Test} to evaluate the trained models on unseen data.

\subsection{Caller-Groups}
\begin{figure}[ht]
  \centering
  \includegraphics[width=\linewidth]{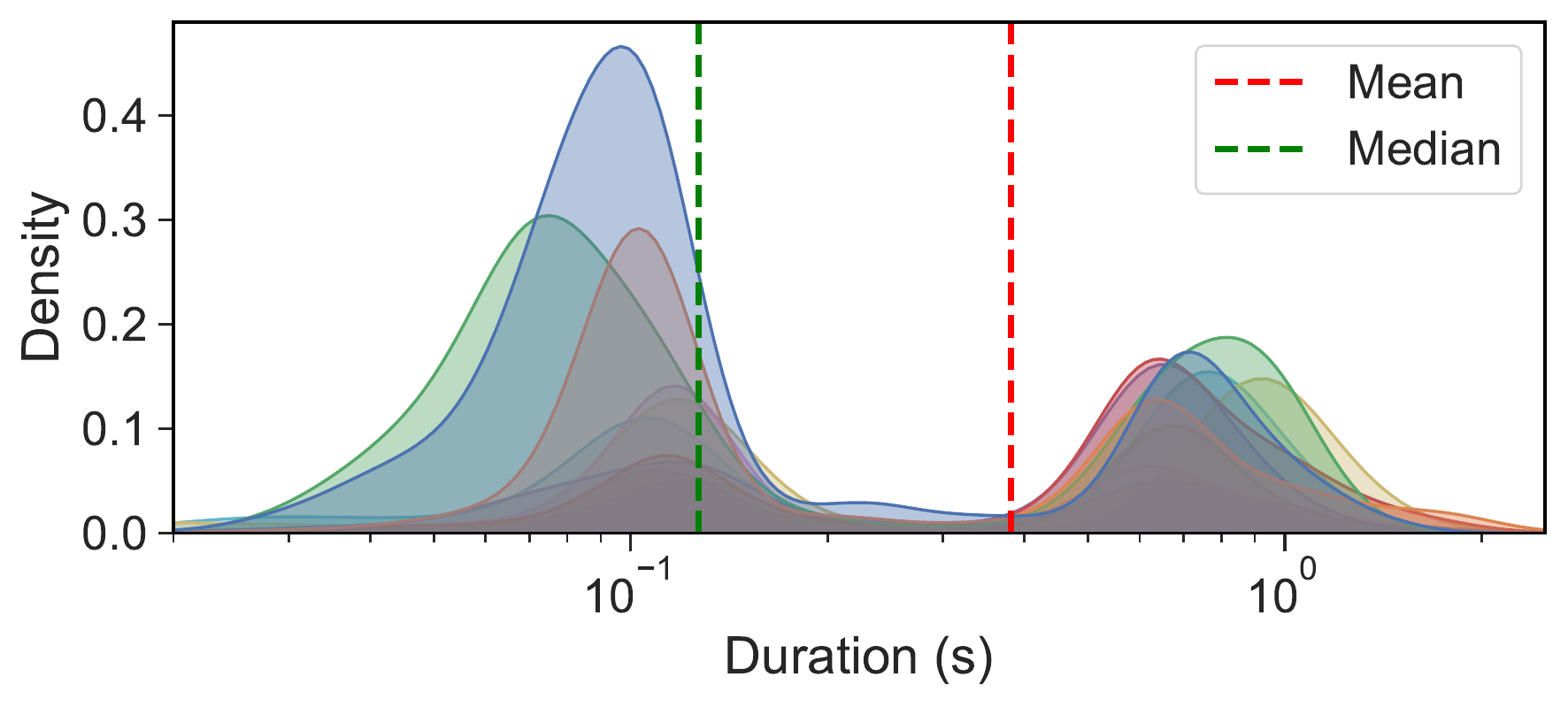}
  \caption{Log distribution of vocalization lengths for callers 1--10 represented in different colors. The mean and median are calculated over the entire dataset.}
  \label{fig:data_duration}
\end{figure}
For our study, neural embeddings are extracted from the pre-trained SSL models by giving the Marmoset vocalizations as input for the purpose of caller detection. The log distribution of vocalization lengths in this dataset, depicted in Figure \ref{fig:data_duration}, exhibits a bimodal structure consistent with prior findings \cite{Huang2022.08.03.502601, TAKAHASHI20132162}. However, the same figure also illustrates that the vocalization segments in this dataset are predominantly short, with a median segment length of around 125 ms. Considering the lack of prior knowledge for this task, we took inspiration from i-vector and x-vector based speaker verification systems, where utterance lengths considerably longer than a short-term window size are modeled to achieve high performance~\cite{5545402, 8461375}. More precisely, in order to effectively model each caller while accounting for the low vocalization segment length as well as to explore the acoustic variations within each caller, we first split all the vocalization embeddings by caller. Then, in order to maintain the chosen 70:20:10 split ratio of our data sets, we divide the embeddings of each caller sequentially into a fixed number of groups, hereafter referred to as `caller-groups'. We set the number of said groups to 100 for \textit{Train}, and proportionally scale for \textit{Val} and \textit{Test}. This results in a total of 1000, 280, and 140 groups across all callers for \textit{Train}, \textit{Val}, and \textit{Test} sets, respectively.

\subsection{Embedding Spaces}
We carry out caller discrimination analysis and caller detection studies by computing the first and second order statistics of the SSL embeddings in the caller-groups. For this purpose, we select eleven pre-trained SSL models from the SUBERB leaderboard \cite{yang21c_interspeech} based on the different pretext tasks seen in \Cref{sec:intro}, and use the S3PRL toolkit \cite{yang21c_interspeech} to extract the embeddings. \Cref{table:ssl_models} lists the chosen models, along with their number of parameters $P$ in millions, and the dimension $D$ of the last layer embedding. All the models have been pre-trained on the LibriSpeech (LS) corpus, except Modified-CPC which is pre-trained on the Libri-Light (LL) corpus.
\begin{table}[ht]
\centering
\caption{Selected pre-trained SSL models on human speech. $P$ indicates the number of parameters in millions, and $D$ corresponds to the dimension of the last layer embedding.}
\begin{tabular}{llrrl}
\toprule
\textbf{Model} & \textbf{Corpus} & $\bm{P}$ & $\bm{D}$ & \textbf{Pretext Obj.} \\
\midrule
APC \cite{chung2019unsupervised}           & LS 360 & 4.11 & 512 & Autoreg. Rec. \\ 
VQ-APC \cite{chung2020vqapc}               & LS 360 & 4.63 & 512 & Autoreg. Rec. \\ 
\midrule
NPC \cite{liu2020nonautoregressive}        & LS 360 & 19.38 & 512 & Masked Rec.\\ 
Mockingjay \cite{9054458}                  & LS 100 & 21.33 & 768 & Masked Rec. \\ 
TERA \cite{10.1109/TASLP.2021.3095662}     & LS 100 & 21.33 & 768 & Masked Rec. \\ 
\midrule
Mod-CPC \cite{riviere2020unsupervised}     & LL 60k & 1.84 & 256 & Contrastive \\ 
Wav2Vec2 \cite{baevski2020wav2vec}         & LS 960 & 95.04 & 768 & Contrastive \\ 
\midrule
Hubert \cite{hsu2021hubert}                & LS 960 & 94.68 & 768 & Masked Pred. \\ 
DistilHubert \cite{chang2022distilhubert}  & LS 960 & 27.03 & 768 & Masked Pred. \\ 
WavLM \cite{9814838}                       & LS 960 & 94.38 & 768 & Masked Pred.\\ 
Data2Vec \cite{pmlr-v162-baevski22a}       & LS 960 & 93.16 & 768 & Masked Pred. \\ 
\bottomrule
\end{tabular}
\label{table:ssl_models}
\end{table}

\section{Caller Discrimination Analysis}
\label{sec:analysis}
This section presents a discrimination analysis of SSL embedding spaces for the purpose of marmoset caller distinction. For this study we only use the \textit{Train} portion of the data.


In order to conduct this analysis on our data, we first model the embedding spaces of each caller-group with a multivariate Gaussian distribution $\mathcal{N}(\bm{\mu}, \bm{\Sigma})$ with mean $\bm{\mu}$ and diagonal covariance matrix $\bm{\Sigma}$, resulting in a total of 100 multivariate Gaussians for each caller.

Subsequently, we compute the inter-caller and intra-caller distances by comparing the multivariate Gaussian distributions. Specifically, for inter-caller distances, we calculate a total of $100 \cdot 100$ pairwise distances for each pair of callers. For intra-caller distances, we compute a total of $\binom{100}{2}$ distances. To compute the distance between the the Gaussians of a pair of caller-groups, we use two measures, namely the Kullback-Leibler (KL) divergence and Bhattacharyya distance, both of which produce distances in the range of $[0,+\infty)$. The latter provides a symmetric measure while the former does not.

\begin{figure}[t]
  \centering
  \includegraphics[width=\linewidth]{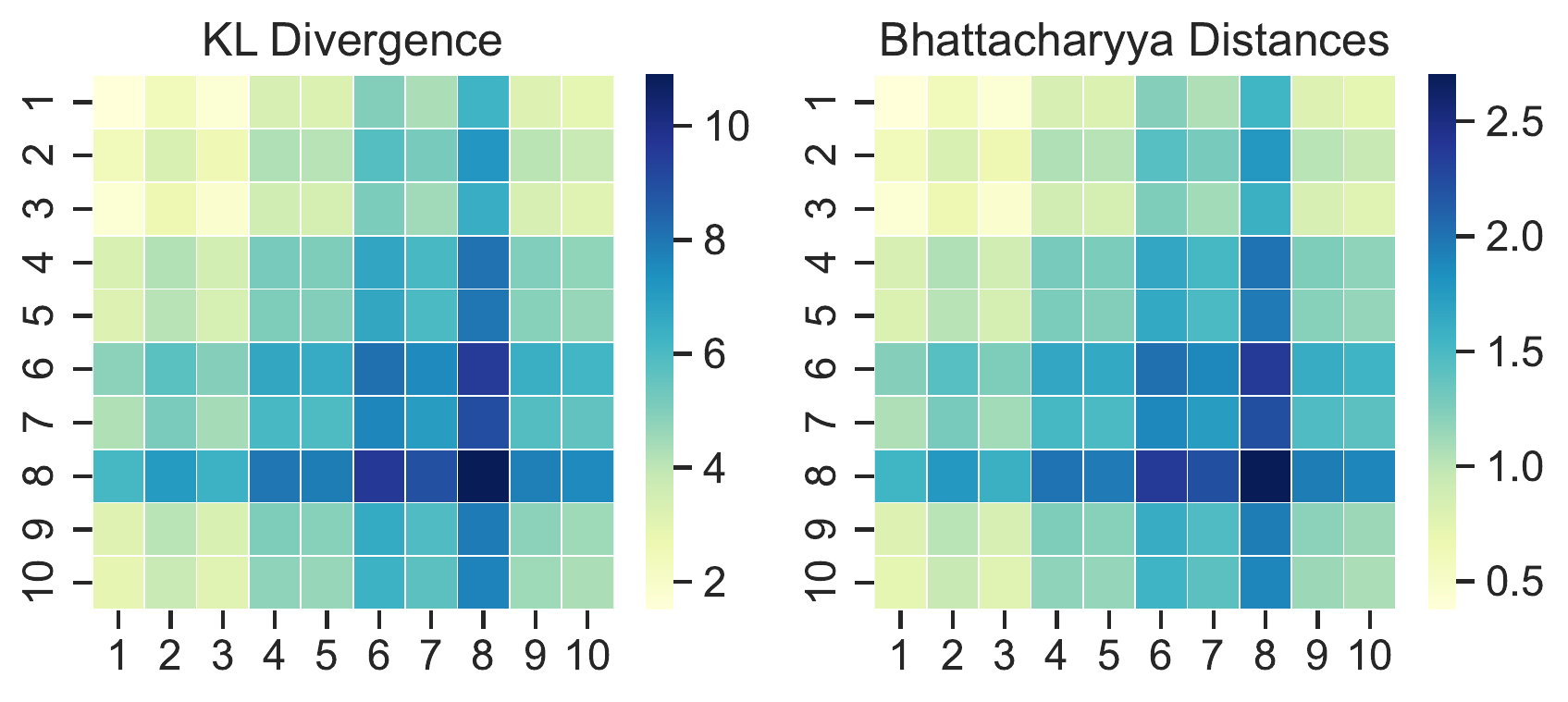}
  \caption{Distance matrix of callers in WavLM's embedding space. The off-diagonal values represent the average inter-caller distances, while the diagonal entries the average intra-caller distances. Darker regions indicate higher dissimilarity.}
  \label{fig:cm}
\end{figure}

Equations \ref{kl_d} and \ref{bc_d} respectively provide the formulas for calculating the KL divergence $D_{\text{KL}}$ and Bhattacharyya distances $D_{BC}$ between two multivariate Gaussian distributions $\mathcal{N}_f$ and $\mathcal{N}_g$ \cite{kl_d, bc_d}. In the case of the KL divergence, the mean vector $\bm{\mu}$, covariance matrix $\bm{\Sigma}$, determinant $\lvert \Sigma \rvert$, and dimensionality $d$ are utilized. Meanwhile, the Bhattacharyya distance uses the arithmetic mean of the covariance matrices $\bm{\Sigma_f}$ and $\bm{\Sigma_g}$ as $\bm{\Sigma}$.
\begin{equation} \label{kl_d}
\begin{split}
    D_{\text{KL}}(f \lvert\rvert g) = &\frac{1}{2} \Bigl( \log{\frac{\lvert \bm{\Sigma_g} \rvert}{\lvert \bm{\Sigma_f} \rvert}} + \text{Tr}(\bm{\Sigma_g}^{-1}\bm{\Sigma_f}) + \\ 
    &(\bm{\mu}_f - \bm{\mu}_g)^T \bm{\Sigma_g}^{-1} (\bm{\mu}_f - \bm{\mu}_g) - d \Bigl)
\end{split}
\end{equation}

\begin{equation} \label{bc_d}
\begin{split}
    D_{BC}(f||g) = &\frac{1}{8} (\bm{\mu}_f - \bm{\mu}_g)^T \bm{\Sigma}^{-1} (\bm{\mu}_f - \bm{\mu}_g) + \\
    & \frac{1}{2} \log (\frac{\lvert \Sigma \rvert}{\sqrt{\lvert \Sigma_f \rvert \lvert \Sigma_g \rvert}})
\end{split}
\end{equation}
Once we have computed the distribution of distances for all the SSL embedding spaces, we can visualize them through a heatmap. \Cref{fig:cm} shows the distance matrix for WavLM's embedding space, where the diagonal entries represent the intra-caller distances and the off-diagonal correspond to the inter-caller distances. In an ideal scenario, one would expect the intra-class distances between distributions to be smaller than the inter-class ones, which is not entirely the case in our results. Nevertheless, for callers with a larger amount of available data, we can observe good discrimination when compared to callers with a lower amount of data, as in the case of Caller 1 and Caller 3 vs. Caller 8. We observe that the distances exhibit similar patterns for all other SSL embeddings, which suggests these embeddings provide similar information for the caller discrimination task. Taken together, the analysis suggests that the SSL embeddings do carry information for distinguishing marmoset callers to a certain extent. However, accomplishing this simple with a linear classifier may be a challenging task.

\section{Caller Detection Study}
\begin{figure*}[!htb]
  \centering
  \includegraphics[width=\linewidth]{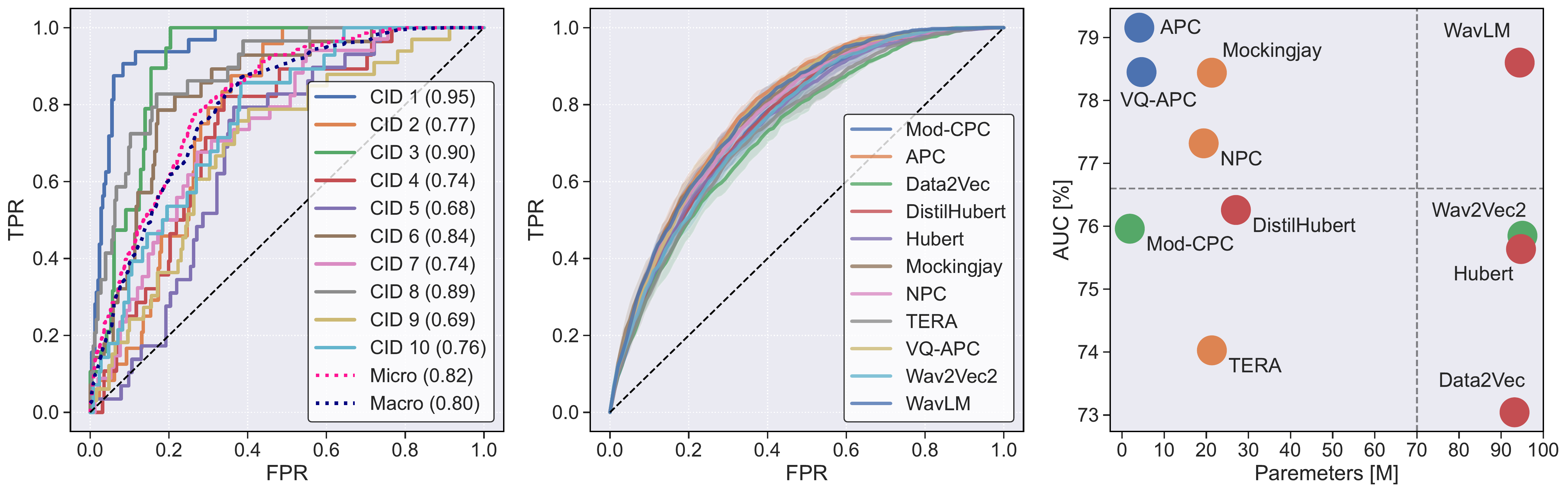}
  \caption{a) ROC curves per caller class (CID) for WavLM embeddings using SVM on one fold of \textit{Test}. b) Macro average ROC curves of all models on \textit{Test} using SVM over all folds. Shaded areas represent $\pm$ 1 std over the k-folds. c) Model size against performance. Model pre-training objective denoted as: \textcolor{sns_red}{$\bullet$} Masked prediction. \textcolor{sns_blue}{$\bullet$} Autoregressive reconstruction. \textcolor{sns_green}{$\bullet$} Contrastive \textcolor{sns_orange}{$\bullet$} Masked reconstruction.} 
  \label{fig:multi_plot}
\end{figure*}

\subsection{Classifiers}
\begin{table}[ht]
\centering
\caption{Search space to find optimal hyperparameters.}
\begin{tabular}{lll}
\toprule
\textbf{Classifier} & \textbf{Hyperparameters} & \textbf{Search space} \\
\midrule
\multirow{4}{*}{RF} & \# Estimators      & [50, 500, 1000, 2000] \\
                     & Max \# Features   & [`auto', `sqrt', `log2'] \\
                     & Criterion         & [`gini', `entropy'] \\
                     & Min samples leaf  & [1, 2, 4] \\
\midrule
\multirow{3}{*}{AB}  & Learning rate     & [0.1, 0.2, 0.5, 1] \\
                     & Algorithms        & [SAMME, SAMME.R] \\
                     & Max \# Estimators & [50, 500, 1000, 2000] \\
\midrule
\multirow{3}{*}{SVM} & C                 & 1e[-5, -4, -3, -2, -1, 0] \\
                     & Kernel            & [RBF, Linear, Polynomial] \\
                     & Gamma             & [`scale', `auto'] \\
\midrule
\multirow{3}{*}{LSVM} & C                 & 1e[-5, -4, -3, -2, -1, 0] \\
                     & Max \# Iterations & 10000 \\
                     & Class weights     & [`balanced', `None'] \\
\bottomrule
\end{tabular}
\label{tab:search_space}
\end{table}
Based on the insights of our caller discrimination analysis, we proceed to classify the statistics computed over the caller-groups for the task of caller detection in a 5 fold cross-validation (CV) framework. We concatenate the mean and variance of the Gaussians into a single functional vector, and use them as our fixed-length representations for classification.

We use Random Forest (RF), Ada Boost (AB), Support Vector Machines (SVM), and Linear SVM (LSVM) algorithms to classify the computed functional vectors. The difference between Linear SVM and SVM with a linear kernel lies in the former's utilization of a squared hinge-loss, while the latter employs a regular hinge-loss.

To determine the most robust classification technique, we employ the grid search methodology with F1-Macro score as the optimization criterion, integrated into the Scikit-learn toolkit. We tune the hyperparameters for each fold, across the train and validation sets over the search space given in \Cref{tab:search_space}.

\subsection{Evaluation Metrics}
To evaluate the effectiveness of our proposed approach for the given task, we present the area under the curve (AUC) scores, which provide a evaluation of the performance of all the classifiers in correctly classifying the positive instances against negative. For SVM it is computed pairwise using a `one-vs-one' methodology, while for the other classifiers it is calculated in a binary `one-vs-rest' framework, by averaging the AUC scores for each class against all others.


\subsection{Results and Discussion} \label{sec:results}
\begin{table}[ht]
\centering
\caption{Macro AUC scores [\%] on Test with 5-fold CV for caller detection task using different classifiers.}
\label{tab:auc_scores}
\begin{tabular}{lrrrr}
\toprule
\textbf{Model} &    \textbf{AB} &  \textbf{LSVM} &    \textbf{RF} &   \textbf{SVM} \\
\midrule
APC          & 71.44 &  65.18 & 70.89 & 79.16 \\
VQ-APC       & 71.60 &  65.58 & 70.04 & 78.45 \\
\midrule
NPC          & 72.61 &  66.27 & 71.50 & 77.32 \\
Mockingjay   & 72.39 &  64.43 & 71.75 & 78.44 \\
TERA         & 70.34 &  64.57 & 68.43 & 74.03 \\
\midrule
Mod-CPC      & 72.62 &  64.05 & 69.81 & 75.96 \\
Wav2Vec2     & 74.41 &  63.94 & 70.18 & 75.85 \\
\midrule
Hubert       & 71.71 &  64.14 & 70.17 & 75.64 \\
DistilHubert & 70.77 &  65.11 & 70.34 & 76.26 \\
WavLM        & 73.97 &  65.32 & 70.74 & 78.60 \\
Data2Vec     & 69.81 &  62.58 & 68.23 & 73.04 \\
\midrule
Average      & 71.97 &	64.66 &	70.19 &	\textbf{76.61} \\
\bottomrule
\end{tabular}
\end{table}
\Cref{tab:auc_scores} summarizes the performance of the different classifiers on all the embedding spaces. The results show that SVM significantly outperforms the other classifiers across all embedding spaces. The decision tree-based ensemble methods, AdaBoost and Random Forest, exhibit comparable performance for most models, and consistently outperform Linear SVM. This suggests that the relationship between the features in the embedding space and their labels is likely to be complex and non-linear, which can be modelled by ensemble methods to some degree, but not to the extent of non-linear SVMs.

\Cref{fig:multi_plot}a) shows the caller classification performance in distinguishing a positive class from the negative instances using SVM on a single \textit{Test} fold. We can observe that all callers are systematically distinguished in this binary framework, including the classes with a low amount of data (CID 6--8). 

\Cref{fig:multi_plot}b) visualizes SVM's average performance for each embedding space across the 5 folds, with the shaded areas representing $\pm$ 1 std. The results clearly demonstrate that the embedding spaces of all models are capable of successfully differentiating Marmoset callers, indicating that SSL models pre-trained on human speech data can generate salient representations capable of distinguishing animal vocalizations regardless of the pre-training criterion. 

\Cref{fig:multi_plot}c) illustrates the relationship between the number of parameters and classification performance for each embedding space. The plot is divided into four quadrants to highlight differences in performance. Interestingly, WavLM's embedding space is found to be more separable than the other masked prediction models, indicating that its masked speech denoising task may be more effective in capturing animal caller identification information than Hubert's masked speech modeling. Surprisingly, both auto-regressive reconstruction based models perform exceptionally well with significantly fewer parameters. These findings suggest that while all pre-training criteria can yield competitive performance, some may be more efficient than others, allowing models with simpler architectures and fewer parameters, such as APC and AQ-APC, to perform comparably to larger models like WavLM. Finally, we observe that Data2Vec is not as successful as the other masked prediction based models, despite the same number of pre-training hours, corpus and comparable number of parameters. While it has shown to outperform the other masked prediction models in human speech, it seems to clearly learn weaker representations for the task of domain adaptation.

\section{Conclusions} \label{sec:conclusion}
This paper investigated the applicability of self-supervised representations, pre-trained on human speech through different approaches, to analyze vocalizations in the bio-acoustics domain. To that end, we conducted and validated two lines of investigation on Marmoset calls in a caller detection framework.


We first conducted a caller discrimination analysis study on the training data to examine the linear separability of eleven pre-trained embedding spaces by splitting the training data into caller-groups, and then calculating the intra-group and inter-group distances through a multivariate Gaussian distribution framework. The results showed that all spaces exhibited similar distance patterns, and that distinguishing marmoset callers is possible with a linear classifier but only to a certain extent.

For our second investigation, we conducted a caller detection study to analyze whether the embedding spaces of said caller-groups can be systematically distinguished by class. We trained four classifiers to predict the classes of the caller-groups in 5 fold cross-validation framework. The results show that we can effectively distinguish all Marmoset callers, including those with low data, in a binary classification framework. The results also show that non-linear SVMs are able to most accurately model the non-linear relationship between the features of the embedding space. Finally, we observe that although all embedding spaces seem effective at the caller detection task, some learning objectives may be more efficient than others.



In summary, our research demonstrates that self-supervised representations pre-trained on human speech can effectively classify vocalizations in the bio-acoustics domain for tasks such as Marmoset caller detection, even without fine-tuning. These findings can greatly benefit bio-acoustics researchers looking to distinguish individual identities within a specific species in their acoustic data. Additionally, we anticipate that further fine-tuning of these models on relevant bio-acoustics downstream tasks can improve performance. Therefore, we plan to investigate the impact of model size on performance after fine-tuning, and also explore adapting the embedding spaces for other tasks like call-type classification in our future work.

\section{Acknowledgments}
This work was funded by Swiss National Science Foundation's NCCR Evolving Language project (grant no. 51NF40\_180888).

\bibliographystyle{IEEEtran}
\bibliography{mybib}

\end{document}